%% file: conference_101719.tex
\documentclass[conference]{IEEEtran}
\IEEEoverridecommandlockouts
\usepackage{cite}
\usepackage{amsmath,amssymb,amsfonts}
\usepackage{algorithmic}
\usepackage{graphicx}
\usepackage{textcomp}
\usepackage{xcolor}
\def\BibTeX{{\rm B\kern-.05em{\sc i\kern-.025em b}\kern-.08em
    T\kern-.1667em\lower.7ex\hbox{E}\kern-.125emX}}
\usepackage{booktabs}
\usepackage{array}
\usepackage{multirow}
\usepackage{url}

\usepackage{xcolor}
\usepackage{cite}

\usepackage{CJKutf8}
\newcommand{\zh}[1]{\begin{CJK}{UTF8}{gkai}#1\end{CJK}} 
\newcommand{\zhT}[1]{\begin{CJK}{UTF8}{bkai}#1\end{CJK}} 

\newcounter{example}

\begin{document}

\title{Towards Comprehensive Semantic Speech Embeddings for Chinese Dialects
}

\author{\IEEEauthorblockN{1\textsuperscript{st} Kalvin Chang*\thanks{*Work completed during an internship at Tencent.}}
\IEEEauthorblockA{\textit{EECS Dept.} \\
\textit{UC Berkeley}\\
Berkeley, U.S.A. \\
kalvinchang@berkeley.edu}
\and
\IEEEauthorblockN{2\textsuperscript{nd} Yiwen Shao}
\IEEEauthorblockA{\textit{Tencent AI Labs} \\
\textit{Tencent}\\
Bellevue, U.S.A. \\
yiwenyshao@global.tencent.com}
\and
\IEEEauthorblockN{3\textsuperscript{rd} Jiahong Li}
\IEEEauthorblockA{\textit{Tencent AI Labs} \\
\textit{Tencent}\\
Shanghai, China}
\and
\IEEEauthorblockN{4\textsuperscript{th} Dong Yu}
\IEEEauthorblockA{\textit{Tencent AI Labs} \\
\textit{Tencent}\\
Bellevue, U.S.A.}
}

\maketitle

\begin{abstract}

Despite having hundreds of millions of speakers, Chinese dialects lag behind Mandarin in speech and language technologies.
Most varieties are primarily spoken, making dialect-to-Mandarin speech-LLMs (large language models) more practical than dialect LLMs.
Building dialect-to-Mandarin speech-LLMs requires speech representations with cross-dialect semantic alignment between Chinese dialects and Mandarin.
In this paper, we achieve such a cross-dialect semantic alignment by training a speech encoder with ASR (automatic speech recognition)-only data, as demonstrated by speech-to-speech retrieval on a new benchmark of spoken Chinese varieties that we contribute.
Our speech encoder further demonstrates state-of-the-art ASR performance on Chinese dialects.
Together, our Chinese dialect benchmark, semantically aligned speech representations, and speech-to-speech retrieval evaluation lay the groundwork for future Chinese dialect speech-LLMs. We release the benchmark at \url{https://github.com/kalvinchang/yubao}.


\end{abstract}


\input{sections/1_intro.tex}

\input{sections/2_methodology.tex}
\input{sections/3_experiments.tex}
\input{sections/5_conclusion.tex}

\section{Acknowledgments}
We acknowledge the efforts of \cite{yubao} in collecting YuBao, without which this work could not exist.
In the spirit of this paper’s focus on preserving linguistic diversity, we close by expressing our gratitude in several Chinese varieties:\\
\zhT{多謝} to-siā (Minnan) / do1 'xia5 (Gan) / do1 sie4 (Xiang) \\
\zhT{恁仔細} an2 zii2 se3 (Hakka) \\
\zhT{唔該} m4 goi1 (Yue) \\
\zhT{謝謝} xiè xie (Mandarin) / 2zia6-zia6 (Wu)

\newpage

\bibliographystyle{IEEEtran}
\bibliography{bibliography}

\end{document}

%% file: sections/1_intro.tex
\section{Introduction}

The performance of speech and language technologies for Chinese dialects still lags behind that of Mandarin, despite the former having 400 million speakers \cite{du2015chinese}. 

\subsection{Background}
Chinese dialects are mutually \textit{un}intelligible.
Thus linguists classify them as distinct languages within the Sinitic family \cite{handel2015}, varieties within a macrolanguage \cite{iso639,ethnologue}, or topolects (regional varieties) \cite{mair1991chinese}. We adopt the term \textit{dialect} hereinafter to be consistent with the ASR (automatic speech recognition) literature.
The mutually \textit{un}intelligibility among Chinese dialects comes from the vast differences in their pronunciation and lexicon \cite{ho2015}.
Despite these differences, Chinese varieties can be classified into a few subgroups: Mandarin, Xiang, Gan, Wu, Min, Hakka, and Yue \cite{handel2015}, and sometimes Jin, Hui, Pinghua, and Tuhua \cite{atlas}).
Building speech and language technologies that support multiple dialect subgroups is inherently a multilingual translation problem. 



\subsection{Motivation}

Our ultimate goal is to build a speech translation model that translates Chinese dialect speech to Mandarin text as input to a large language model (LLM), to broaden access to LLMs to speakers of Chinese dialects, thus encouraging use of Chinese dialects. 
A speech translation model would be more useful than a Chinese dialect ASR-LLM pipeline, since most speakers do not write Chinese dialects.
Even when Chinese dialects are written, the orthography is not standardized for all dialects \cite{ueda2024creating,lau-etal-2025-data}.
Within the same dialect, different scholars or speakers may arrive at different Chinese characters to represent dialect characters \cite{cockrum2023reanalyzing}.
In informal settings, speakers can use phonetically similar but semantically different Chinese characters to transcribe Chinese dialect speech \cite{adhoc-han,khoo2019dynamics}.
This phonetic approximation may even occur in data annotation.
For instance, in one of our training corpora, a Shanghainese Wu utterance is transcribed as:

\begin{center}
\zh{欧一生下来就啥个才会。}
(\refstepcounter{example}\label{ex:shanghai}\theexample)
\end{center}

where \zh{欧} is used to approximate the Shanghainese Wu pronunciation of \zh{我}, the first-person singular pronoun, despite \zh{欧} not having such a meaning.
In short, Chinese dialect ASR is not our ultimate goal because differences in dialect transcription prevent meaningful comparison across datasets or even annotators; instead, we focus on text-free speech-to-speech retrieval.

The first step in building a Chinese dialect-to-Mandarin speech-LLM is to build a speech encoder that maps utterances from different Chinese dialect subgroups, including Mandarin, into a shared semantic space.
The obvious solution would be to train a dialect-to-Mandarin speech translation model.
However, paired dialect speech-Mandarin text (speech translation data) is not as abundant as ASR data.
For many dialect subgroups, the problem necessarily becomes zero-shot speech translation (to Mandarin).
Even without paired ST data, we show that it is still possible to induce a cross-dialect semantic space.

We learn speech representations where semantically similar words and sentences have similar embeddings, demonstrating cross-lingual (semantic) alignment \cite{hammerl-etal-2024-understanding}.
The degree of such alignment can be measured with retrieval, a task that tests whether representations from a source language utterance can be matched to representations of a semantically equivalent utterance in a target language \cite{hammerl-etal-2024-understanding}.
\cite{ma-etal-2025-cross} in particular used speech-to-speech retrieval on FLEURS \cite{conneau2023fleurs} to show that Whisper's speech encoder has a cross-lingual semantic space, which holds even after removing confounders like cognates (shared words across related languages) and named entities that might have inflated the retrieval scores \cite{shim2025languagesmultilingualspeechfoundation}.
Thus in this paper, we leverage speech-to-speech retrieval to measure the degree of cross-dialect alignment in our speech representation space.\footnote{We chose speech-to-speech retrieval over speech-to-text retrieval because strong comprehensive dialect text embeddings currently do not exist.}

Towards this end, we introduce YuBao, a new dataset of parallel speech across Chinese dialects with comprehensive coverage of Chinese dialect subgroups.
Evaluated on speech-to-speech retrieval using our new dataset YuBao, our speech encoder demonstrates strong cross-dialect semantic alignment.
Our speech-to-speech retrieval benchmark provides an additional evaluation for future work on Chinese dialect speech-LLMs.\footnote{We release the benchmark at \url{https://github.com/kalvinchang/yubao}}

\subsection{Related Work}



Prior work on Chinese dialect ASR built speech-LLMs \cite{bai2024seed, xu2025fireredasr, xu2025leveraging, li2025baichuan}, self-supervised speech models \cite{telespeechpt},
attention encoder-decoder models \cite{xu2025fireredasr,radford2022whisper,ding2024chinese}, 
mixture-of-expert models \cite{jie-etal-2024-dialectmoe}.
With the exception of Seed-ASR \cite{bai2024seed} and TeleSpeech's SSL model \cite{telespeechpt}, prior work does not comprehensively cover major Chinese subgroups, 
often focusing on Mandarin dialects only \cite{tang2021kespeech,li2025baichuan,xu2025fireredasr,shen2024multi},
two or more dialects \cite{jie-etal-2024-dialectmoe,xu2025leveraging,dan2022multi,ding2024chinese,chen2024towards,zhang2022chinese,radford2022whisper,feng2025voxlect},
or a single dialect \cite{chou2023evaluating,liao2023taiwanese,liao2022taiwanese,yu-etal-2022-automatic,li2025jlms25,xu-etal-2018-building,radford2022whisper}.
\cite{li2024chinese}'s survey of Chinese dialect ASR highlights the need for an ASR model with comprehensive coverage of Chinese dialects.
Unlike related work, we achieve state-of-the-art Chinese dialect ASR with a Zipformer encoder \cite{yao2023zipformer} and a non-LLM attention decoder. Our model also comprehensively covers most major Chinese dialect subgroups (minus Jin and Gan).

Within prior work on Chinese dialect speech translation, there is translation from Hokkien to Mandarin \cite{liao2020,liang-2021,lin2024minspeech}, from Cantonese to English \cite{xiao23d_interspeech}, and Taiwanese Hokkien to English \cite{meta-s2st}, but not multi-dialect translation to Mandarin.
Our speech representation space with cross-dialect semantic alignment makes significant strides towards multi-dialect translation to Mandarin.

\subsection{Contribution}

In short, we contribute the following:
\begin{itemize}
    \item YuBao, a new Chinese dialect speech dataset with comprehensive coverage of Chinese dialect subgroups (Sec~\ref{sec:yubao})
    \item A state-of-the-art dialect ASR model with comprehensive coverage of Chinese dialect subgroups (Sec~\ref{sec:asr-model})
    \item Speech representations with cross-dialect semantic alignment induced with ASR-only data, measured by cross-dialect speech-to-speech retrieval (Sec~\ref{sec:retrieval})
\end{itemize}

%% file: sections/2_methodology.tex
\section{Methodology}

\subsection{YuBao: a New Chinese Dialect Speech Benchmark}
\label{sec:yubao}

Our Chinese dialect speech retrieval benchmark comes from the Centre for the Protection of Language Resources of China \cite{yubao}, which we abbreviate as YuBao (\zhT{語保}).
The original YuBao website has dialect speech, dialect transcripts, IPA transcripts, and Mandarin translations for 1,000 characters, 1,200 words, and 50 sentences, all of which are parallel (semantically aligned), across 1,300+ sites in China.
For our retrieval benchmark, we leveraged the up to 50 parallel sentences available (some sites had slightly less than 50), which consists of read speech from older males \cite{yubao}, who are more likely to preserve more linguistic features of their dialects than younger speakers.
We chose 11 sites for each subgroup of Chinese/Sinitic and additionally scraped 1 site, Luanping, to represent Standard Mandarin, 
for a total of 78 sites, spanning the seven major subgroups: Mandarin (dialectal Mandarin), Yue, Min (Southern Min/Minnan), Hakka, Xiang, Wu, and Gan.
We focused on sites where the subgrouping info was provided by YuBao or sites where the subgrouping is unambiguous (e.g. Changsha belonging to Xiang).
The Mandarin spoken in Luanping, Hebei---not Beijing---is the closest to Standard Mandarin, phonetically speaking \cite{luanping}.
Overall, our retrieval benchmark consists of 3,499 utterances on average $6.9 \pm 3.0$ seconds long, for a total of 6 hours, 45 minutes.

\subsection{Dialect ASR with Comprehensive Coverage of Sinitic}
\label{sec:asr-model}

Our Chinese dialect ASR model was trained on 34,000 hours of data with coverage of most major Chinese subgroups: Mandarin, Yue, Wu, Min, Hakka, and Xiang (Tab.~\ref{tab:asr-train}).
All data is clean and has a 16k Hz sampling rate. The Chinese dialect speech mostly contains spontaneous and read speech, except for conversational Hakka. The proprietary dialectal Mandarin and accented Mandarin data contains speech from non-standard varieties of Mandarin and Mandarin spoken by speakers whose native dialect is not Mandarin. This data includes conversational speech from Dongbei (Northeastern) Mandarin and spontaneous speech from the following cities or regions across China: Zhengzhou, Lanzhou, Ningxia, Shanxi, Liaoning, Yunnan, Hefei, Sichuan, Tianjin, Guilin, Wuhan, Jiangxi, Hebei, Jinan, Zhejiang, Jiangsu, Anhui, Hunan, Fujian, Xi'an, and Changsha.
We did not modify noisy text transcriptions (as highlighted in Ex.~\ref{ex:shanghai}) as this is the result of a lack of a standard orthography, and we did not have the resources to re-annotate the entire corpus.
Moving forward, we concur with \cite{lau-etal-2025-data}'s  call for more enforcing orthographic standards in the transcription process.

\begin{figure}
    \centering
    \includegraphics[width=\linewidth]{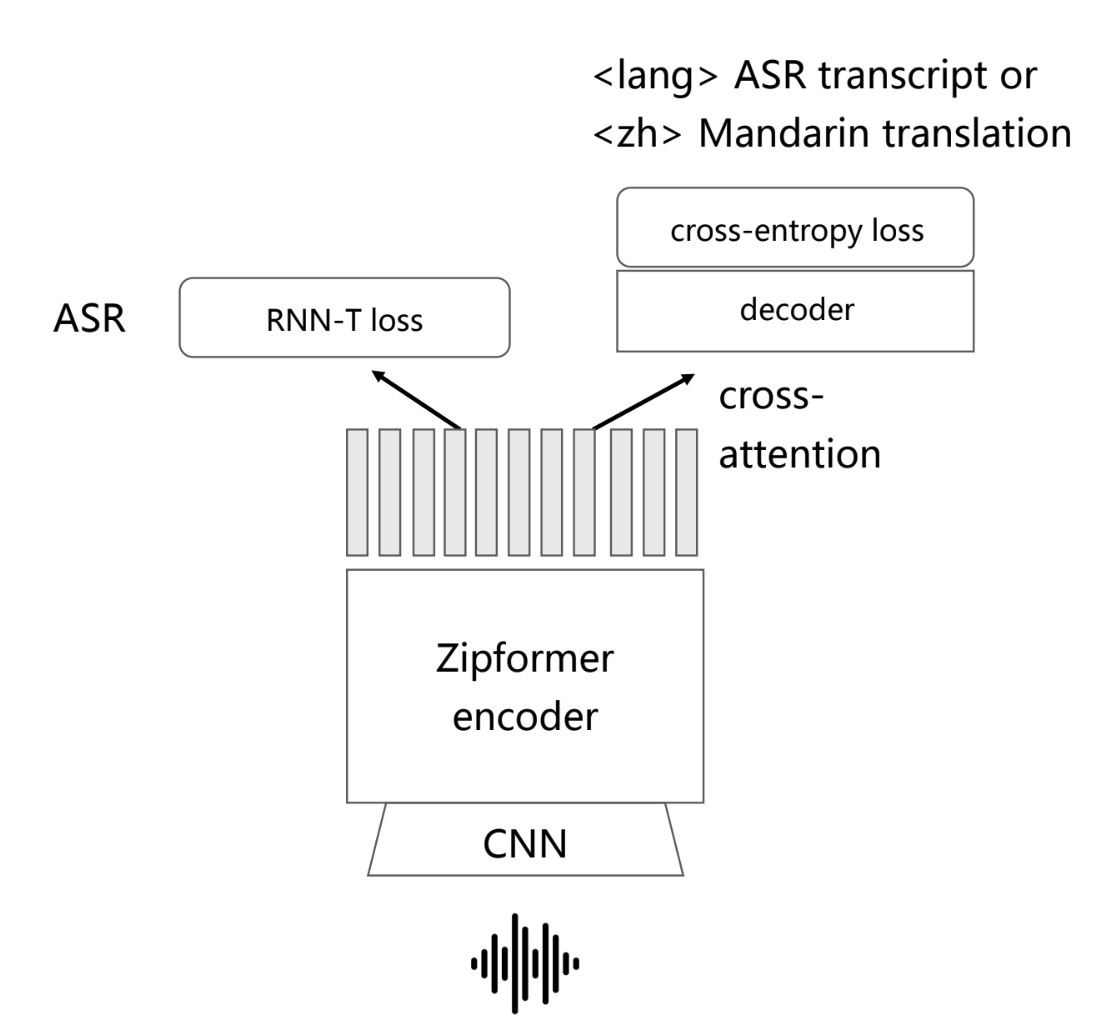}
    \caption{Architecture of our models trained with ASR-only or ASR + speech translation (ST) data (Sec.~\ref{sec:asr-model}). During training, ASR data goes through both the RNN-T head and the attention head (``decoder''), while ST goes through the attention head (``decoder'') only.
    During inference, the RNN-T head is used for ASR, while the attention head (``decoder'') can be used for ST. The gray boxes illustrate the speech encoder embeddings used in our retrieval experiments.}
    \label{fig:model-architecture}
\end{figure}

We used a Zipformer encoder \cite{yao2023zipformer}, which achieved state-of-art ASR performance while being faster and less memory-intensive than the Conformer \cite{gulati20_interspeech} due to its U-Net-like downsampling towards the middle.
The Zipformer encoder was trained with a joint pruned \cite{kuang2022pruned} RNN-T loss \cite{graves2012sequence} and attention loss \cite{watanabe2017hybrid}, where label smoothing \cite{szegedy2016rethinking} was used for the attention loss.
We use the RNN-T head (joiner, decoder) for ASR decoding and the attention head (commonly referred to as the decoder) for translation (Sec.~\ref{sec:st-experiment}).
See Fig.~\ref{fig:model-architecture} for a diagram of the model.
The model contains 186,806,275 parameters, with 19 Zipformer encoder layers and 6 Transformer decoder layers.
The decoder is intentionally weak to concentrate semantic abilities in the encoder.
To perform ASR inference, greedy decoding was applied to the encoder only for the data in Tab.~\ref{tab:asr-test}.
We used \cite{yao2023zipformer}'s ScaledAdam optimizer and Eden learning rate scheduler to train our model, which they showed to converge more quickly than and perform better than Adam \cite{kingma2014adam}.

\begin{table}
\center
\caption{Dialect ASR and speech translation training data
}
\label{tab:asr-train}
\begin{tabular}{>{\raggedright\arraybackslash}p{1cm}ccc}
\toprule
Subgroup & Corpus & Hours & Labels \\
\midrule
Mandarin & WeNetSpeech \cite{zhang2022wenetspeech} & 10,005 & ASR \\
& KeSpeech \cite{tang2021kespeech} & 1,396 & ASR \\
& AISHELL-1 \cite{bu2017aishell}, AISHELL-2 \cite{du2018aishell} & 1,150 & ASR \\
& Mandarin dialects+accents across China & 11,151 & ASR \\
Yue & Cantonese & 4,528 & ASR \\ 
Wu & Shanghai & 525 & ASR \\
& Shanghai2 & 493 & ASR \\
& Hangzhou & 234 & ASR \\
& Suzhou & 162 & ASR \\
Min & Chaoshan & 352 & ASR \\
& Hokkien & 968 & ASR \\
& Hokkien2 & 141 & ASR \\
Hakka & Meizhou & 494 & ASR, ST\\
Xiang & Changsha & 984 & ASR, ST \\
\midrule
English & LibriSpeech \cite{panayotov2015librispeech} & 960 & ASR \\
En-Zh Codeswitching & TALCS \cite{li22j_interspeech} & 555 & ASR \\
\midrule
Total & All & 34,098 & - \\
\bottomrule
\end{tabular}
\end{table}

\begin{table}
\center
\caption{Dialect ASR test set}
\label{tab:asr-test}
\begin{tabular}{llcc}
\toprule
Subgroup & Corpus & Hours\\
\midrule
Mandarin & AISHELL-1 \cite{bu2017aishell}, AISHELL-2 \cite{bu2017aishell} & 10 \\
& KeSpeech \cite{tang2021kespeech} & 31\\
Yue & CV-yue \cite{ardila-etal-2020-common} & 15 min \\
& MDCC \cite{yu-etal-2022-automatic}  & 11 \\
Wu & Shanghai & 9.5 \\
& Shanghai2  & 9.5 \\
& Hangzhou & 15 \\
& Suzhou & 14 \\
Min & Chaoshan & 9 \\
& Hokkien & 10 \\
& Hokkien2 & 10.5 \\
Xiang & Changsha & 10 \\
\bottomrule
\end{tabular}
\end{table}

\subsection{Zero-shot Speech-to-speech Retrieval}
\label{sec:retrieval}

\begin{figure*}
    \centering
    \includegraphics[width=0.6\linewidth]{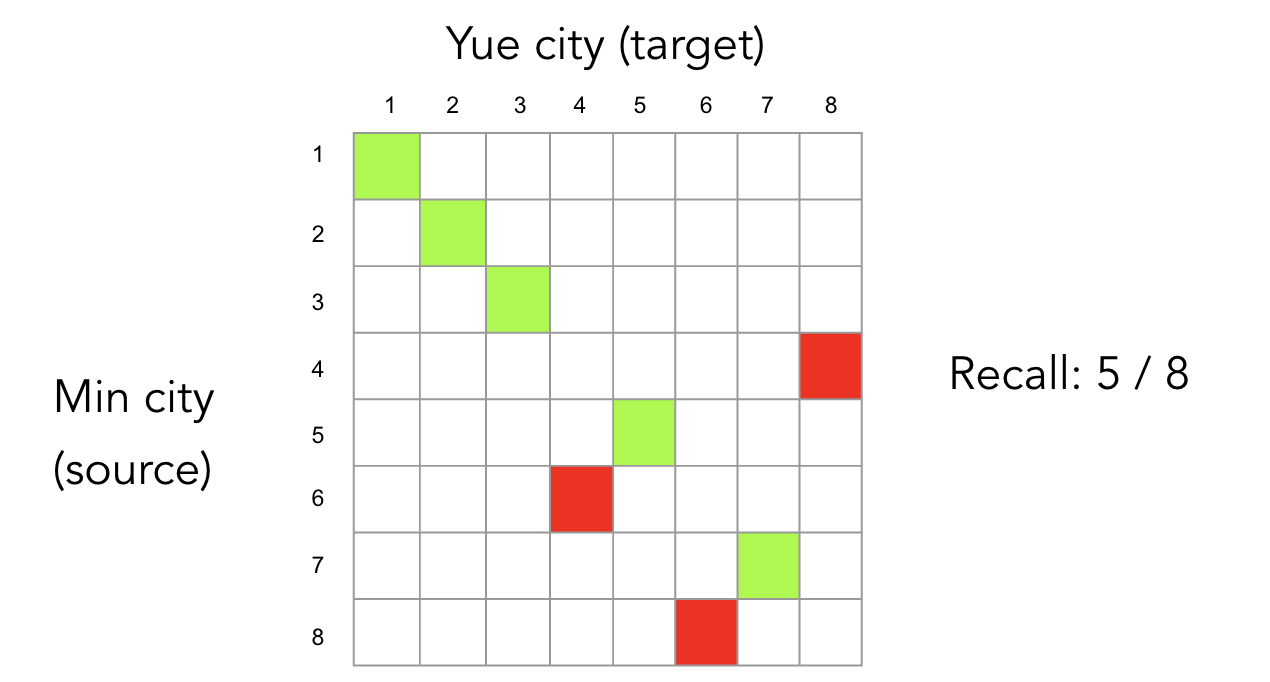}
    \caption{Illustration of speech-to-speech retrieval between a pair of dialect sites. Suppose there is a spoken corpus composed of 8 sentences with the same meaning across different dialects. Then the goal is to measure how well the speech embeddings can match the utterance in a source dialect to the utterance in the target dialect with the same meaning. The matching (retrieval) is done by identifying the target dialect utterance with the highest embedding similarity (Fig.~\ref{fig:seq-sim}) to the source dialect utterance.
    Embedding similarity (Fig.~\ref{fig:seq-sim}) is computed for each each cell in this figure, which represents one pair of sentences between a source and target dialect.
    A retrieved pair is identified as correct if both utterances have the same meaning, i.e. they lie along the diagonal in this figure.
    }
    \label{fig:retrieval}
\end{figure*}

We evaluated the cross-dialect semantic alignment of our speech encoder with recall on the task of zero-shot speech-to-speech retrieval \cite{ma-etal-2025-cross,zanon-boito-etal-2020-mass,duquenne2023sonar},
using our YuBao benchmark, which consists of 50 parallel sentences from YuBao across 78 sites (Sec~\ref{sec:yubao}).
Given the speech representations of an utterance in a source language, we retrieve the utterance in a target language with the highest embedding similarity to the source utterance's representations.
If the retrieved utterance in the target language has the same meaning as the source utterance, then the retrieved utterance is correct (see Fig.~\ref{fig:retrieval}).
In other words, for each sentence in the source site, does the sentence in the target site with the highest speech embedding similarity have the same meaning as the source dialect?
This is measured by the recall rate between a pair of cities.
We compute the recall between all pairs of cities within a source and target subgroup and report the mean.

\begin{figure*}
    \centering
    \includegraphics[width=\linewidth]{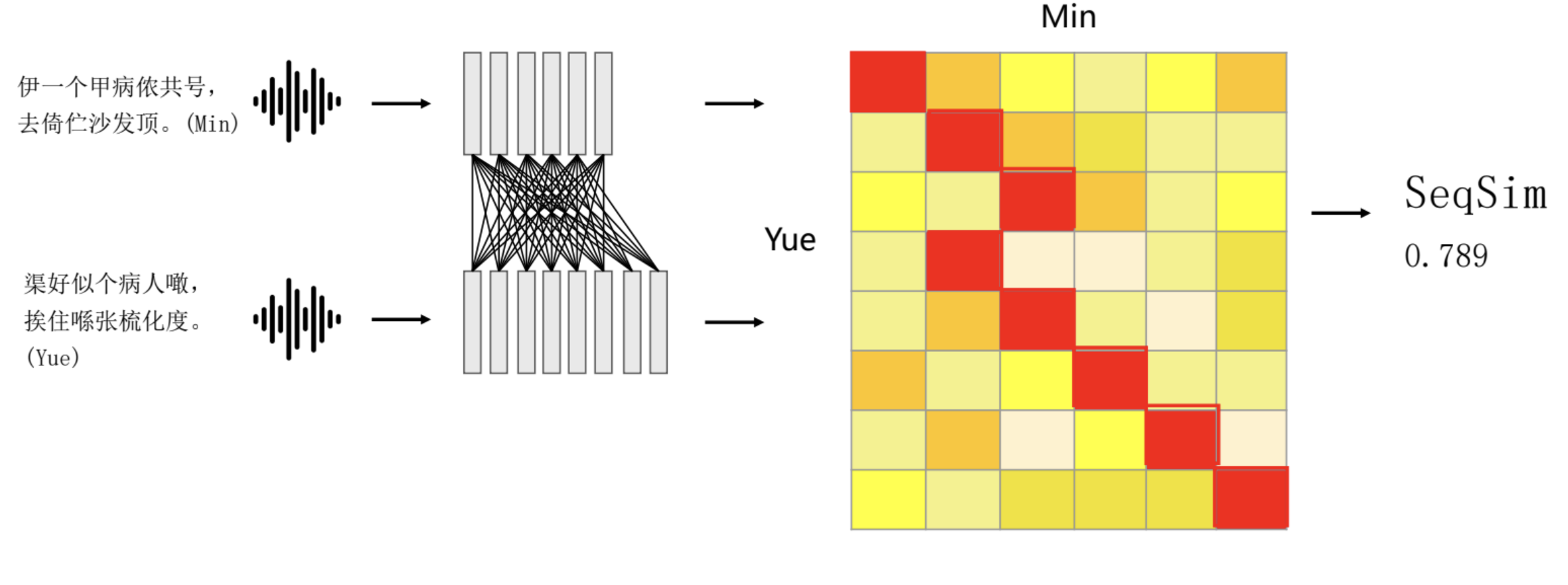}
    \caption{Illustration of how SeqSim \cite{ma-etal-2025-cross} between a pair of Chinese dialect sites in YuBao is computed. The cosine similarity between all pairs of speech encoder frames between a source utterance's embeddings and a target utterance's embeddings is calculated. Then the maximum similarity is taken across each row and column to obtain the final SeqSim score according to \eqref{eq:seqsim}.
    }
    \label{fig:seq-sim}
\end{figure*}

The embedding similarity between the representations of a source and a target utterance is measured with SeqSim, proposed by \cite{ma-etal-2025-cross}.
SeqSim is essentially a frame-level BERTScore \cite{zhangbertscore}, where for all source-target pairs of time steps (tokens in NLP, frames in speech), we take the cosine similarity between representations at the pair of time steps:
\begin{align}
    \text{Re}_{\text{seq}} &= \dfrac{1}{|X|} \sum_{\mathbf{x} \in X} \max_{\mathbf{y} \in Y} \boldsymbol{x}^\mathsf{T} \boldsymbol{y} \\
    \text{Pr}_{\text{seq}} &= \dfrac{1}{|Y|} \sum_{\mathbf{y} \in Y} \max_{\mathbf{x} \in X} \boldsymbol{x}^\mathsf{T} \boldsymbol{y} \\
    \text{SeqSim} &= 2 \cdot \dfrac{\text{Pr}_{\text{seq}} \cdot \text{Re}_{\text{seq}}}{\text{Pr}_{\text{seq}} + \text{Re}_{\text{seq}}}
    \label{eq:seqsim}
\end{align}
where $\textbf{x, y} \in \mathbb{R}^{D}$ (single encoder frames), $X \in \mathbb{R}^{T_1 \times D}, Y \in \mathbb{R}^{T_2 \times D}$ (encoder embeddings). See Fig.~\ref{fig:seq-sim} for an illustration.

If our encoder maps Chinese dialects to a shared semantic space, then the speech-to-speech retrieval recall rate should be above random.

%% file: sections/3_experiments.tex
\section{Experiments}

\subsection{Dialect ASR}

Models were trained with icefall \cite{icefall}.
Data processing and speech feature extraction were performed with lhotse \cite{zelasko2021lhotse}.
We used FBANK features with 80 mel bins and applied SpecAugment \cite{park19e_interspeech}.
We applied text normalization to both the reference and ASR hypotheses during evaluation on the test set, converting traditional to simplified Chinese, removing \textit{erhua}, and introducing spaces between Chinese characters.
Each model was trained for 2 weeks using 32 GPUs.
See Tab.~\ref{tab:asr-hyperparams} for model and optimization hyperparameters.

As shown in Tab.~\ref{tab:asr-results}, our Zipformer-based ASR model outperforms the Paraformer \cite{gao22b_interspeech} and FireRed-AED \cite{xu2025fireredasr} models for almost all Chinese varieties, except for Mandarin.
The Paraformer is considered a strong (Standard) Mandarin model but was never trained on dialects.
Similarly, FireRed-AED was trained on Standard Mandarin and Mandarin dialects but not others.
Despite this, we compared our models with Paraformer and FireRed AED because they are also attention encoder-decoder models, as opposed to Speech-LLMs.
Another reason for choosing these two models was that ASR models trained with comprehensive coverage of Chinese dialects, such as \cite{bai2024seed}, are not public. As for \cite{telespeechpt}, only the SSL data had comprehensive coverage of Chinese dialect subgroups; they did release an ASR checkpoint, but it was only finetuned on Mandarin dialects \cite{tang2021kespeech}.
Additionally, \cite{xu2025leveraging}, which has significantly larger decoders than our models, is not public.

Surprisingly, Paraformer and FireRed-AED perform better on the Changsha corpus than on other non-Mandarin test sets.
We hypothesize this is because Changsha Xiang belongs to the New Xiang subcluster, which is similar to and even somewhat intelligible with Southwestern Mandarin due to long-term dialect contact \cite{you2025chinese}.


\begin{table}[h]
\centering
\caption{ASR Model Hyperparameters}
\label{tab:asr-hyperparams}
\begin{tabular}{ll}
\toprule
\textbf{Hyperparameter}         & \textbf{Value}                        \\
\midrule
\textit{Model:} & \\
feature\_dim                   & 80                                   \\
output\_downsampling\_factor  & 2                                    \\
num\_encoder\_layers           & {[}2, 2, 4, 5, 4, 2{]}               \\
downsampling\_factor           & {[}1, 2, 4, 8, 4, 2{]}               \\
encoder\_dim                   & {[}192, 256, 512, 768, 512, 256{]}   \\
feedforward\_dim               & {[}576, 768, 1536, 2304, 1536, 768{]}\\
dropout                       & null                                 \\
num\_heads                    & {[}4, 4, 4, 8, 4, 4{]}               \\
query\_head\_dim              & {[}32{]}                            \\
value\_head\_dim              & {[}12{]}                            \\
pos\_head\_dim                & {[}4{]}                             \\
pos\_dim                      & 48                                   \\
encoder\_unmasked\_dim        & {[}192, 192, 256, 256, 256, 192{]}   \\
cnn\_module\_kernel           & {[}31, 31, 15, 15, 15, 31{]}         \\
decoder\_dim                  & 512                                  \\
joiner\_dim                   & 512                                  \\
causal                        & false                                \\
chunk\_size                   & {[}16, 32, 64, -1{]}                 \\
left\_context\_frames         & {[}64, 128, 256, -1{]}               \\
special\_tokens               & null                                 \\
vocab\_size                   & 2000                                 \\
blank\_id                     & 0                                    \\
context\_size                 & 2                                    \\
max\_duration                 & 400 seconds                          \\
\midrule
\textit{Optimizer:} & \\
base\_lr                      & 0.045 \\
lr\_epochs                     & 3.5 \\
lr\_batches                    & 7500 \\
lr\_steps\_per\_epoch         & 100000                              \\
warmup\_batches                & 4000                              \\
lm\_scale                & 0.25                              \\
am\_scale                & 0                              \\
simple\_loss\_scale        & 0.5 \\
rnnt\_warm\_step           & 2000 \\
use\_fp16 & True \\
attention\_loss\_scale & 0.8 \\
balance\_loss\_scale & 0 \\
specialization\_loss\_scale & 0 \\
\bottomrule
\end{tabular}
\label{tab:hyperparams}
\end{table}

\subsection{ST}
\label{sec:st-experiment}

\cite{ma-etal-2025-cross} improved the performance of X-to-Mandarin speech translation by finetuning Whisper on English-to-Mandarin speech translation.
This suggests that finetuning on speech translation strengthens the cross-lingual alignment already present in Whisper (shown by speech-to-speech retrieval prior to finetuning), even for languages not represented in the speech translation data.
We thus sought whether limited speech translation data can enhance the cross-dialect semantic retrieval.
We trained an additional Zipformer model using the 1478 hours of Hakka and Xiang ST data we had (Tab.~\ref{tab:asr-train}).
The hyperparameters of the ASR+ST model are the same as the model trained with ASR-only data (Tab.~\ref{tab:asr-hyperparams}).
Since \cite{peng23d_interspeech,ma-etal-2025-cross} showed that speech translation with the ASR task token (with the target language token) performs better than translation with the ST task token, we do not use any task token and simply use the Mandarin language token to perform speech translation to Mandarin using the decoder.
ASR data goes through both the RNN-T and the attention head, while ST data only goes through the attention head with the target language token as a prefix.
See Fig.~\ref{fig:model-architecture} for a diagram of the model.
(The ASR-only model has the same architecture but was not trained on ST data.)
The model trained on ASR+ST data achieved similar ASR performance to the model trained only on ASR data (Tab.~\ref{tab:asr-results}).

\subsection{Retrieval}

Our retrieval results, shown in Tab.~\ref{tab:retrieval-asr-st-enc-dec} and Tab.~\ref{tab:retrieval-asr-only-enc-dec}, suggest a cross-dialect shared space emerges, even with ASR-only data.
Unsurprisingly, the similarity between Standard Mandarin and dialectal Mandarin---members of the same dialect subgroup---is high in both directions.
Furthermore, the Mandarin-dialect and dialect-Mandarin retrieval recall rates are almost all above 80\%, with the exception of Gan, which did not appear in our training data.
This demonstrates that the dialects share a common semantic space with Mandarin, which suggests that our speech encoder can be used with a Mandarin LLM.
Specifically, retrieving the correct Mandarin sentence for a dialect sentence suggests that a Speech-LLM is likely to understand the same sentence.
Additionally, the retrieval recall is greatly above random chance between all pairs of dialect subgroups for both of our models, including between Standard Mandarin and other subgroups (in both directions).
This indicates that both our models learn a cross-dialect semantic space between all pairs of dialect subgroups, not just between Mandarin and the dialects.

Additionally, that the model trained with ASR-only data (Tab.~\ref{tab:retrieval-asr-only-enc-dec}) demonstrates similar retrieval recall to the model trained with both ASR and ST data (Tab.~\ref{tab:retrieval-asr-st-enc-dec}) suggests that ASR-only data---not ST data---is sufficient to learn a cross-dialect semantic space.
This is surprising because \cite{shim2025languagesmultilingualspeechfoundation} argue that speech translation is the contributor behind cross-lingual semantic alignment in speech-to-text foundation models.


We hypothesize that cross-dialect semantic alignment arises from ASR-only data because of semantic supervision from text transcripts.
Language modeling in text learns a form of distributional semantics \cite{mikolov-etal-2013-linguistic}, and 
multilingual language models in particular can learn cross-lingual semantics even without parallel data \cite{pires-etal-2019-multilingual,conneau2019cross}.
Furthermore, using ASR data is known to strengthen semantics when learning speech-text alignment \cite{huzaifah2023analysis,arora2025landscape}.
This corroborates \cite{shim2025languagesmultilingualspeechfoundation}'s finding that OWSM v3.1 Small Low-Restriction \cite{peng24b_interspeech}, trained with ASR-only data, demonstrates cross-lingual semantic retrieval capabilities between related languages. 
In short, supervision from text transcripts in our 34,000 hours of paired cross-dialect speech and text imbues the encoder with a cross-dialect semantic space.



\begin{table}[!h]
\center
\caption{Character error rate for our Zipformer models trained on ASR+ST (320k steps) and ASR only data (312k steps), decoded with greedy search applied to the encoder only}
\label{tab:asr-results}
\resizebox {\linewidth} {!} {
\begin{tabular}{ll|cc|cc}
\toprule
& & \multicolumn{2}{c}{ours} & \multicolumn{2}{c}{theirs} \\
Subgroup & Corpus & ASR+ST & ASR-only & Paraformer & FireRed-AED \\
\midrule
Mandarin & AISHELL & 1.69 & 1.85 & 1.87 & 0.54 \\
& KeSpeech & 5.45 & 5.75 & 11.26 & 4.85 \\
Yue & CV-yue & 7.44 & 7.48 & 73.23 & 45.79 \\
& MDCC & 6.37 & 6.28 & 47.72 & 36.18 \\
Wu & Shanghai & 10.02 & 10.25 & 66.42 & 58.34 \\
& Shanghai2 & 7.36 & 7.54 & 66.70 & 56.14 \\
& Hangzhou & 4.95 & 5.25 & 46.10 & 37.73 \\
& Suzhou & 10.17 & 10.46 & 80.62 & 77.56 \\
Min & Chaoshan & 10.60 & 10.79 & 83.59 & 78.64 \\
& Hokkien & 21.34 & 21.66 & 90.63 & 88.65 \\
& Hokkien2 & 21.00 & 22.01 & 92.19 & 89.20 \\
Xiang & Changsha & 6.21 & 7.83 & 33.94 & 28.71 \\
\bottomrule
\end{tabular}
}
\end{table}

\begin{table}[!t]
\center
\caption{Speech-to-speech retrieval recall rates for Zipformer model trained with ASR and ST data\\ (320k steps)}
\label{tab:retrieval-asr-st-enc-dec}
\resizebox {\linewidth} {!} {
\begin{tabular}{lccccccc}
\toprule
Source \textbackslash Target & Mandarin (Std) & Mandarin (Dialect) & Min & Wu & Yue & Xiang & Gan \\
\midrule
Mandarin (Std) & -- & 98.2 & 93.8 & 89.3 & 83.6 & 91.8 & 72.7 \\
Mandarin (Dialect)  & 99.6 & -- & 88.6 & 85.8 & 79.8 & 90.0 & 70.8 \\
Min & 95.3 & 90.0 & -- & 85.3 & 80.8 & 87.5 & 69.2 \\
Wu & 95.3 & 89.6 & 86.1 & -- & 75.7 & 87.8 & 69.5 \\
Yue & 87.5 & 80.0 & 79.6 & 72.7 & -- & 78.3 & 62.1 \\
Xiang & 94.4 & 89.2 & 85.8 & 81.3 & 74.1 & -- & 71.5 \\
Gan & 78.0 & 72.4 & 67.1 & 64.8 & 59.0 & 73.1 & -- \\
\bottomrule
\end{tabular}
}
\end{table}

\begin{table}[!h]
\center
\caption{Speech-to-speech retrieval recall rates for Zipformer model trained with ASR data only\\ (312k steps)
}
\label{tab:retrieval-asr-only-enc-dec}
\resizebox {\linewidth} {!} {
\begin{tabular}{lccccccc}
\toprule
Source \textbackslash Target & Mandarin (Std) & Mandarin (Dialect) & Min & Wu & Yue & Xiang & Gan \\
\midrule
Mandarin (Std) & -- & 98.9 & 93.1 & 91.3 & 84.9 & 91.5 & 73.1 \\
Mandarin (Dialect) & 99.3 & -- & 89.1 & 86.4 & 80.0 & 89.2 & 70.8 \\
Min & 94.5 & 90.0 & -- & 86.4 & 80.5 & 87.1 & 68.4 \\
Wu & 93.5 & 88.7 & 87.4 & -- & 74.9 & 87.1 & 68.9 \\
Yue & 85.6 & 78.7 & 79.9 & 72.2 & -- & 76.7 & 61.2 \\
Xiang & 91.5 & 88.7 & 85.6 & 81.1 & 75.3 & -- & 71.2 \\
Gan & 77.1 & 71.8 & 67.2 & 64.4 & 58.8 & 72.6 & -- \\
\bottomrule
\end{tabular}
}
\end{table}

%% file: sections/5_conclusion.tex
\section{Conclusion and Future Work}

Our work has achieved state-of-the-art Chinese dialect ASR performance across a comprehensive set of major Chinese dialect subgroups using a Zipformer encoder.
When evaluated on zero-shot speech-to-speech retrieval using our new YuBao benchmark---which has comprehensive coverage of major Chinese dialect subgroups---our encoder representations demonstrate strong cross-dialect semantic alignment.
Our retrieval is particularly beneficial in settings like Chinese dialects without standardized orthographies.
We further showed that such semantically aligned embeddings can be learned without dialect-to-Mandarin ST data or data in scarce dialect-dialect pairs, using ASR-only data.
While we focused on Chinese dialects, our approach can be applied to other closely related language continua with mid- or high-resource varieties, such as Bantu or Indic.
It remains to be seen if this holds true for phylogenetically \textit{un}related languages.



In the future, we hope to expand the YuBao benchmark to all 1300 cities so that we can measure the zero-shot generalization of the cross-dialect alignment to subclusters such as Eastern Min (Mindong) not represented during training.
We will also strengthen the encoder's cross-dialect semantic alignment while still only using ASR-only data via teacher-student distillation  \cite{khurana2022samu,duquenne2023sonar} via contrastive learning \cite{rao2025whispa,srinivasa2023cwcl} where a noisy cross-dialect text model (e.g. SentenceBERT or a Mandarin LLM finetuned on Chinese dialect ASR transcripts)---assumed to have cross-dialect semantic alignment---will teach the student speech model to push utterances with a similar meaning together in the speech representation space.
We also seek to make the Zipformer encoder even more efficient using Mixture-of-Experts \cite{jie-etal-2024-dialectmoe}.
Finally, we seek to create a language family tree or clustering of Chinese dialects \cite{bartelds2022quantifying,shim-nerbonne-2022-dialectr} using SeqSim \cite{ma-etal-2025-cross}, to be compared with linguists' hypotheses about how the Chinese dialects evolved from Old and Middle Chinese \cite{huang2024geographic}.